# MATRIX TILE ANALYSIS

**Inmar Givoni, Vincent Cheung, Brendan J. Frey**
Probabilistic and Statistical Inference Group
University of Toronto
10 King's College Road, Toronto, Ontario, Canada, M5S 3G4

## Abstract

Many tasks require finding groups of elements in a matrix of numbers, symbols or class likelihoods. One approach is to use efficient bi- or tri-linear factorization techniques including PCA, ICA, sparse matrix factorization and plaid analysis. These techniques are not appropriate when addition and multiplication of matrix elements are not sensibly defined. More directly, methods like biclustering can be used to classify matrix elements, but these methods make the overly-restrictive assumption that the class of each element is a function of a row class and a column class. We introduce a general computational problem, 'matrix tile analysis' (MTA), which consists of decomposing a matrix into a set of non-overlapping tiles, each of which is defined by a subset of usually nonadjacent rows and columns. MTA does not require an algebra for combining tiles, but must search over an exponential number of discrete combinations of tile assignments. We describe a loopy BP (sum-product) algorithm and an ICM algorithm for performing MTA. We compare the effectiveness of these methods to PCA and the plaid method on hundreds of randomly generated tasks. Using double-gene-knockout data, we show that MTA finds groups of interacting yeast genes that have biologically-related functions.

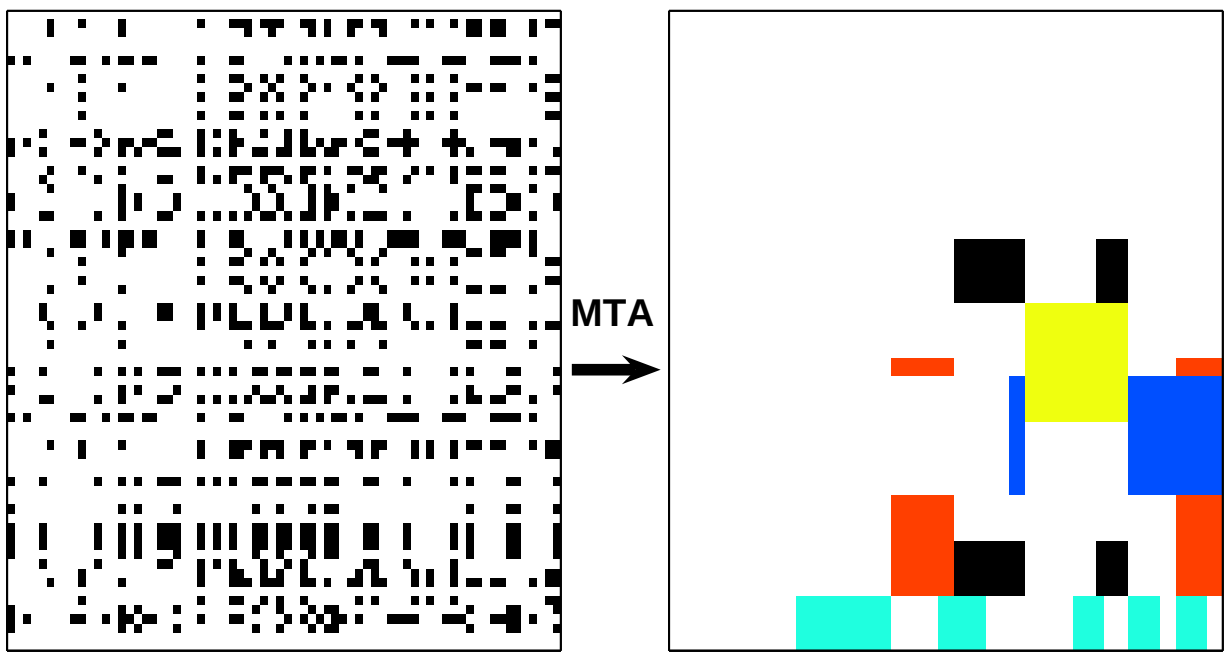

Figure 1: To illustrate matrix tile analysis, we show a binary data matrix (left), and after applying MTA, the same matrix as a collection of tiles (right), having reordered rows and columns to better reveal the tiles. Each tile is described by a subset of rows and columns. Different tiles may not overlap. Each element is colored according to its tile index.

## 1 INTRODUCTION

A variety of data types are most naturally represented as matrices of numbers, symbols, or, after preprocessing, class likelihoods. For example, the viability of a yeast strain obtained by knocking out two genes can be compared to normal viability to obtain an 'abnormal growth likelihood' associated with the pair of genes. Many such double-gene-knockout experiments can be used to construct a matrix of likelihoods (Tong *et al.* 2004). In collaborative filtering (Resnick *et al.* 1994) the matrix rows correspond to clients and the columns correspond to, say, movies that clients are potentially interested in watching. An observed element of the data matrix contains information regarding a preference indicated by the corresponding client for the corresponding movie.

Two approaches to analyzing an $N \times M$ data matrix $X$ include matrix factorization methods and combined row-column clustering techniques. Matrix factorization methods are only appropriate when a sensible algebra (addition and multiplication) can be defined for the data elements. If $X$ contains real numbers, regular addition and multiplication are assumed to be sensible, often without convincing justification. If $X$ contains non-ordinal discrete variables or likelihood functions, there often does not exist a sensible algebra for the elements. Assuming an algebra can be defined, matrix factorization methods find factorizations of the form $X \approx USV$, where $S$ is a diagonal matrix, $U$ is an $N \times J$ matrix and $V$ is a $J \times M$ matrix. Different factorization methods enforce different types of constraints on the matrices such as orthonormality of $U$'s columns and $V$'s rows (PCA, Jolliffe 1986), non-negativity of $U, S, V$ (non-negative matrix factorization, Lee & Seung 1999), and allowing $J$, the size of the subspace the data is projected onto, to be large while requiring sparse solutions (ICA, Bell & Sejnowski 1995, sparse matrix factorization, Dueck *et al.* 2005) or low-norm solutions (maximum-margin matrix factorization, Srebro *et al.* 2005).

Combined row-column clustering techniques classify data elements, instead of finding algebraic decompositions. One approach is to independently cluster the rows and columns (Eisen *et al.* 1999), but this ignores dependencies between the two clustering problems. In bi-clustering (Cheng & Church 2000), each column has a class label and each row has a class label so that the class of each data element is given by joint row-column class labels. So, if there are $n_c$ possible classes for the columns and $n_r$ possible classes for the rows, each element can be in one of $n_c n_r$ classes. While this approach can account for dependencies in the row clustering and column clustering problems, it makes the overly-restrictive assumption that the class of each element is a function of a row class and a column class, resulting in clusters defined by common row and column boundaries. Row classes whose associated elements show similar interaction patterns to a subset of column classes are not grouped, though they correspond to a block of data composed of several identified smaller blocks. This restriction also applies to stochastic block models and their extensions, used mainly for the analysis of relational data (Airoldi *et al.* 2005).

Another approach to combined row-column clustering is to apply a matrix factorization technique, and then binarize the output. After applying PCA, the elements of $U$ and $V$ can be binarized by applying a threshold and the binary patterns in $U$ and $V$ can be used to define classes. The plaid model (Lazzeroni & Owen 2002) finds a decomposition of the same form $X \approx USV$ described above, but the elements of $U$ and $V$ are regularized toward the values 0 or 1 during optimization. While the method of quantizing the output of a matrix factorization technique provides an often computationally efficient solution to joint row-column clustering, selection of thresholds is not straightforward and the technique does not directly optimize a cluster model.

We introduce a new computational approach that we call 'matrix tile analysis' (MTA), which explains the data as a collection of non-overlapping tiles. We formulate MTA as a probabilistic model and present two different algorithms that approximately maximize the joint probability as well as two standard algorithms (the plaid method and PCA) which can be modified to find tiles, and compare the performance of all methods on synthetic data. We use MTA and hierarchical agglomerative clustering to find biologically relevant groupings in the yeast genetic interaction data, and show that MTA finds more biologically informative groups, thus demonstrating the applicability and usefulness of this method.

## 2 MATRIX TILE ANALYSIS

In MTA, an $N \times M$ data matrix $X$ is modelled as a set of non-overlapping regions, or tiles. The order of the rows and columns that is most appropriate for revealing a specific tile by grouping its elements into a contiguous rectangular block may be inappropriate for another tile. Thus, visually each tile is associated with a different permutation of the rows and columns that groups its elements. For example, the binary data in Fig. 1 is rearranged and colored to expose the tile structure of the data. Only one tile is fully contiguous for this particular permutation. The data elements in each tile are accounted for by class-conditional probabilities and data that does not appear in any tile is accounted for by a background model. Unlike bi-clustering, MTA does not require that the tiles are defined by a common set of row and column boundaries. In fact, MTA allows an arbitrary subset of rows and columns to be assigned to each tile, subject to the constraint that two tiles cannot own the same elements. Like matrix factorization techniques, MTA can be thought of as factorizing the input matrix into multiple components or tiles. However, because every element belongs to only one tile, it is not required that a sensible algebra be defined to combine elements from different tiles. In this way, MTA can be applied when sensible notions of addition and multiplication are not available. Finally, MTA can be formulated as a problem of finding tiles, given tile-conditional likelihoods as input, *i.e.*, the analysis method does not need the original data as input.

In a model with $T$ tiles, we denote the index of a particular tile by $t$. We use binary random variables $r_i^t$ to indicate the rows of matrix elements belonging to the tile, *i.e.*, $r_i^t = 1$ indicates that the $i$th row of $X$ contains elements belonging to tile $t$, and $r_i^t = 0$ indicates that none of the elements in the $i$th row of $X$ belong to tile $t$, where $i = 1 \dots N$. Similarly, binary random variables $c_j^t$ indicate those columns containing elements belonging to tile $t$, where $j = 1 \dots M$. The outer product $(r_1^t, r_2^t, \dots, r_N^t)^\top (c_1^t, c_2^t, \dots, c_M^t)$ is an $N \times M$ binary matrix indicating those elements belonging to tile $t$. In total, the hidden variables describing the tile analysis include $T$, $R = \{r_i^t \mid i = 1 \ldots N, t = 1 \ldots T\}$ and $C = \{c_j^t \mid j = 1 \ldots M, t = 1 \ldots T\}$.

Each tile has a user-defined likelihood specifying the probability of membership of elements in $X$ to that tile. We denote the $N \times M$ matrix of likelihoods corresponding to tile $t$ by $L^t$. That is, $\ell_{ij}^t = P(x_{ij} | \text{tile } t)$. To account for elements not placed in tiles, we use $L^0$ to denote the likelihood matrix corresponding to the background model. These likelihood matrices are the input to MTA and are pre-computed depending on the task at hand. For example, if the data matrix is binary-valued and has symmetric noise with probability $\epsilon$, then $\ell_{ij}^t = (1-\epsilon)^{x_{ij}} \epsilon^{1-x_{ij}}$. In this paper, we address the computational task of identifying tiles, assuming the input likelihood matrices can be pre-computed. In the rest of the paper we address the case of a single likelihood model for all tiles, though our model easily extends to the general case of $t$ likelihood matrices.

The overall data likelihood is

$$P(X|R, C, T) = \prod_{i=1}^{N} \prod_{j=1}^{M} (\ell_{ij})^{\sum_{t=1}^{T} r_i^t c_j^t} (\ell_{ij}^0)^{(1-\sum_{t=1}^{T} r_i^t c_j^t)}. \quad (1)$$

We assume that given $T$, $R$ and $C$ are uniformly distributed over all non-overlapping tile configurations[1]:

$$P(R, C|T) = \frac{1}{h(T)} \prod_{i=1}^{N} \prod_{j=1}^{M} \Big[ 0 \leq \sum_{t=1}^{T} r_i^t c_j^t \leq 1 \Big]. \quad (2)$$

where $h(T)$ is used to normalize the distribution. The minimum number of tiles is $T = 0$ and the maximum number of tiles is $T = NM$, in which case there is one tile for every data element. The distribution of $T$ can be specified depending on the application, but in this paper, we assume a uniform distribution: $P(T) = 1/(NM+1), T \in \{0, \ldots, NM\}$.

The probability of the number of tiles, the rows and columns in every tile, and the data is

$$P(X, R, C, T) = P(T)P(R, C|T)P(X|R, C, T). \quad (3)$$

The goal of MTA is to infer the values of $T$, $R$ and $C$, *i.e.*, compute $\text{argmax}_{T,R,C} P(X, R, C, T)$. We describe four different approaches for inferring $R$ and $C$, where $T$ is automatically selected using an MDL framework.

[1]Square brackets are used in the fashion of Iverson's notation, where $[true] = 1$ and $[false] = 0$

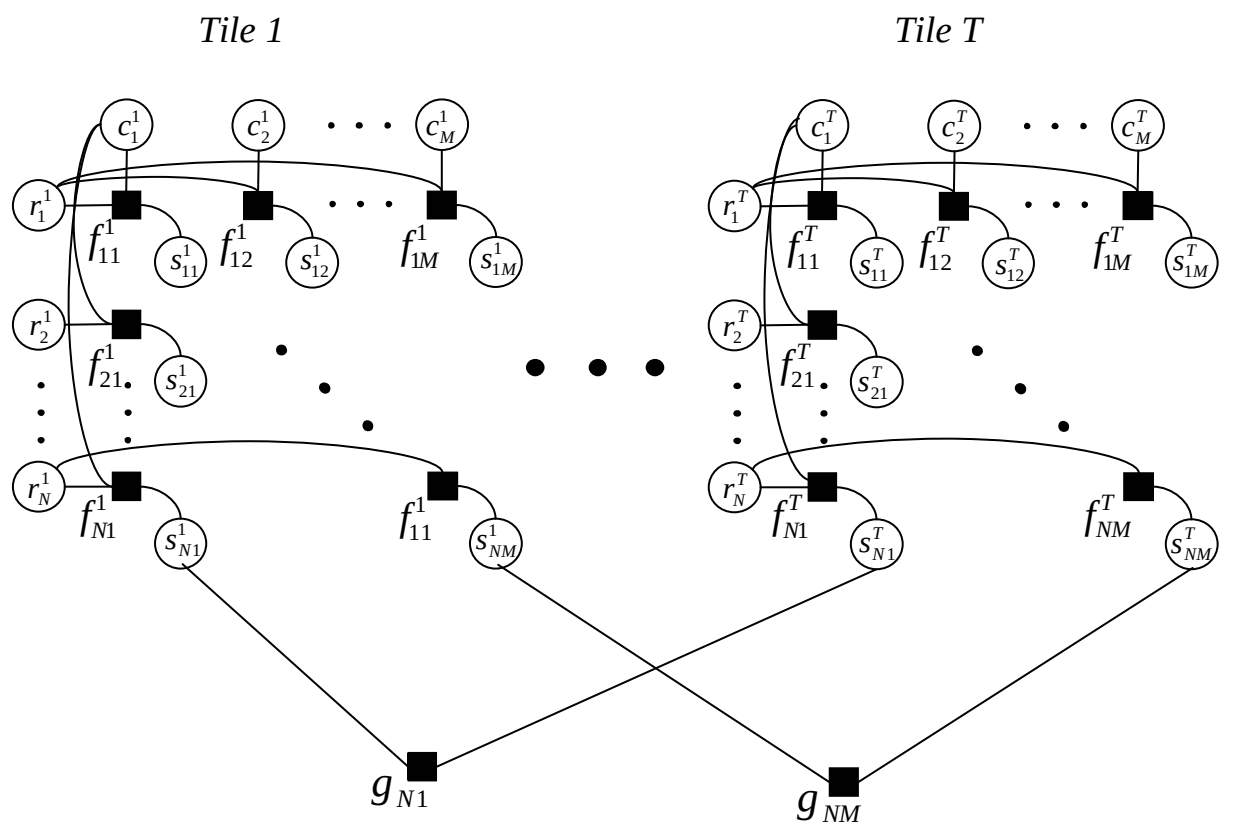

Figure 2: Factor graph for MTA

# 3 MTA INFERENCE

## 3.1 SUM-PRODUCT ALGORITHM

Based on the data model detailed in equations (1-3), we describe a probabilistic graphical model for the matrix tile analysis problem. The model is shown in Fig. 2, using the standard notation for factor graphs (Kschischang *et al.* 2001).

In order for the model to fully describe the constraints requiring tile elements to be non-overlapping, we introduce an additional set of binary random variables, $s_{ij}^t$ whose purpose is to restrict each matrix element to belong to one tile at the most.

The $s_{ij}^t$ nodes indicate whether matrix element $x_{ij}$ is accounted for by tile $t$. The $r_i^t$ nodes are binary random variables that indicate whether row $i$ of the data matrix contains elements which belong to tile $t$. Likewise, node $c_j^t$ is an indicator which is active when column $j$ of the data matrix contains elements belonging to tile $t$.

Each triplet of nodes $(s_{ij}^t, r_i^t, c_j^t)$ is connected to a function node denoted by $f_{ij}^t$, and the function calculated at the node is defined as

$$f_{ij}^t(s_{ij}^t, r_i^t, c_j^t) = \quad (4)$$
$$\begin{cases} 1 & \text{if } (s_{ij}^t = r_i^t = c_j^t = 1) \\ & \text{or } (s_{ij}^t = 0 \text{ and } (r_i^t = 0 \text{ or } c_j^t = 0)); \\ 0 & \text{otherwise.} \end{cases}$$

This function node connects elements within the same tile, and enforces the constraint that if a matrix element $x_{ij}$ is accounted for by tile $t$ (as indicated by variable $s_{ij}^t$ having a value of 1), then the corresponding row and column indicators must be active. If an element $x_{ij}$ is *not* accounted for by tile $t$, then either the row indicator, or the column indicator, or both must be inactive.

For a specific matrix element $x_{ij}$, the corresponding $s$-

variables $s_{ij}^1, s_{ij}^2, \ldots, s_{ij}^T$ are connected to the function node $g_{ij}$ (shown in Fig. 2 only for matrix element $x_{N1}$ and $x_{NM}$ for clarity). This function bridges together $s$-variable nodes from different tiles to enforce the constraint that each matrix element $x_{ij}$ is accounted for by a single tile at most, as well as to account for the likelihood of $x_{ij}$.

$$g_{ij}(s_{ij}^1, s_{ij}^2, \ldots, s_{ij}^T) = \begin{cases} \ell_{ij} & \text{if } s_{ij}^t = 1, s_{ij}^k = 0 \ \forall k \neq t; \\ \ell_{ij}^0 & \text{if } s_{ij}^t = 0 \ \forall t; \\ 0 & \text{otherwise,} \end{cases} \tag{5}$$

where $t = 1 \ldots T, k = 1 \ldots T$. When the constraint is satisfied, so that only one tile accounts for the element, the function evaluates to the likelihood of that element under the model. If no tile accounts for the element, the function evaluates to the likelihood of the background model.

The overall data likelihood as represented in the factor graph is

$$P(X|T,R,C,S) = \prod_{i=1}^{N}\prod_{j=1}^{M}\Big(g_{ij}(s_{ij}^1,..,s_{ij}^T)\prod_{t=1}^{T} f_{ij}^t(s_{ij}^t, r_i^t, c_j^t)\Big), \tag{6}$$

where $S = \{s_{ij}^t \mid i = 1 \ldots N, j = 1 \ldots M, t = 1 \ldots T\}$.

In order to solve the MTA problem we infer $(R, C, S)$ for a set of $T$ tiles, by applying the sum-product algorithm (SP-MTA) on the factor graph shown in Fig. 2.

Messages between nodes are sent as log probability ratios, as they contain all information needed to reconstruct the probability of the binary variable nodes, and they present an efficient implementation which is also robust to extreme probability swings[2].

Denoting by $\mu_{f \to x}$ a message sent from a function node $f$ to a variable node $x$ and $\mu_{x \to f}$ as the message sent from a variable node to a function node, we obtain the following updates for propagation:

$$\begin{aligned}
\mu_{g_{ij} \to s_{ij}^t} &= -\log\Big(\frac{\ell_{ij}^0}{\ell_{ij}} + \sum_{k \neq t} \exp\{\mu_{f_{ij}^k \to s_{ij}^k}\}\Big) \\
\mu_{f_{ij}^t \to r_i^t} &= \log\Big(\frac{\exp\{\mu_{c_j^t \to f_{ij}^t} + \mu_{g_{ij} \to s_{ij}^t}\} + 1}{\exp\{\mu_{c_j^t \to f_{ij}^t}\} + 1}\Big) \\
\mu_{r_i^t \to f_{ij}^t} &= \sum_{k \neq j} \mu_{f_{ik}^t \to r_i^t} \\
\mu_{f_{ij}^t \to c_j^t} &= \log\Big(\frac{\exp\{\mu_{r_i^t \to f_{ij}^t} + \mu_{g_{ij} \to s_{ij}^t}\} + 1}{\exp\{\mu_{r_i^t \to f_{ij}^t}\} + 1}\Big) \\
\mu_{c_j^t \to f_{ij}^t} &= \sum_{k \neq i} \mu_{f_{kj}^t \to c_j^t} \\
\mu_{f_{ij}^t \to s_{ij}^t} &= \mu_{c_j^t \to f_{ij}^t} + \mu_{r_i^t \to f_{ij}^t} - \\
&\quad \log\big(\exp\{\mu_{c_j^t \to f_{ij}^t}\} + \exp\{\mu_{r_i^t \to f_{ij}^t}\} + 1\big),
\end{aligned} \tag{7}$$

and we use the fact that $\mu_{s_{ij}^t \to f_{ij}^t} = \mu_{g_{ij} \to s_{ij}^t}$ and $\mu_{s_{ij}^t \to g_{ij}} = \mu_{f_{ij}^t \to s_{ij}^t}$.

Since the factor graph described in Fig. 2 contains many cycles, the algorithm is used within the loopy belief propagation framework (LBP). There is no known criteria to distinguish among instances of LBP which converge and those who do not. However, there are many examples of well known problems for which LBP converges, and empirically we observe it is also the case for MTA.

Upon termination, we infer $R$ and $C$ by fusing all the incoming messages $\mu_{f_{it}^t \to r_i^t}$ and $\mu_{f_{tj}^t \to c_j^t}$ at each $r_i^t$ and $c_j^t$ nodes respectively, and applying a threshold to obtain the desired binary values. Convergence was determined by evaluating the relative change in the values of the messages arriving from the tile element nodes $s_{ij}^t$ to function nodes $g_{ij}$ after each of the possible messages has been calculated and sent once.[3]

Message scheduling and initialization was found to be important for the algorithm to find good solutions. The messages are processed for each tile consecutively. Within each tile, messages are sent according to the order they are presented in (7). All messages were initialized to be zero in the log domain, except for the messages from matrix element nodes $s_{ij}^t$ to the external function nodes $g_{ij}$, which were the first messages to be propagated, and were set to $-\infty$, indicating that initially, matrix elements are not claimed by any tile. Incremental thresholding of the messages was also found to aid in convergence to stable solutions.

[2] Letting the message $\pi$ be equal to $\frac{\log p(x=1)}{\log p(x=0)}$ for some random variable x, we reconstruct the two probabilities from the message as $p(x=1) = \frac{\exp \pi}{1+\exp \pi}$ and $p(x=0) = \frac{1}{1+\exp \pi}$)

[3] Convergence was declared if $\sum_{i,j,t} |\mu^k_{f_{ij}^t \to s_{ij}^t} - \mu^{k+1}_{f_{ij}^t \to s_{ij}^t}| < 1 \times 10^{-3} \cdot \sum_{i,j,t} |\mu^k_{f_{ij}^t \to s_{ij}^t}|$, where k indicates the iteration.

### 3.2 ICM

An approximate algorithm for maximizing the joint probability in (3) is Iterative Conditional Modes (MTA-ICM). To maximize $P(X, R, C, T)$, or equivalently, $\log P(X, R, C, T)$, we alternate between updating the rows and columns of the tiles, leading to an iterative algorithm with the following updates:

$$\{r_i^1, \ldots, r_i^T\} \leftarrow \operatorname{argmax}_{\{r_i^1,\ldots,r_i^T\}} \Big( \sum_{t,j} r_i^t c_j^t (\log \frac{\ell_{ij}}{\ell_{ij}^0}) + \sum_j \log \Big[ 0 \le \sum_t r_i^t c_j^t = 1 \Big] \Big) \quad (8)$$

$$\{c_i^1, \ldots, c_i^T\} \leftarrow \operatorname{argmax}_{\{c_i^1,\ldots,c_i^T\}} \Big( \sum_{t,i} c_j^t r_i^t (\log \frac{\ell_{ij}}{\ell_{ij}^0}) + \sum_i \log \Big[ 0 \le \sum_t r_i^t c_j^t = 1 \Big] \Big). \quad (9)$$

The constraints are enforced via the last sums in (8) and (9). The "gain" over the base model by including row $i$ in tile $t$ can be defined as $g_i^t = \sum_j c_j^t \big(\log(\ell_{ij}) - \log(\ell_{ij}^0)\big)$. For row $i$, the gain can be computed for each tile independently and the joint probability is optimized for choosing binary values for $\{r_i^1, \ldots, r_i^T\}$ that both maximize the sum of the gains and satisfy the constraints. While only tiles with positive gains need to be considered, an exhaustive search still requires exponential time over the number of tiles with positive gains. The search can be reduced by only searching over configurations that satisfy the constraints.

Let $D$ be a $T \times T$ matrix, where $D_{uv}$ is 1 if row $i$ can be in both tiles $u$ and $v$ ($r_i^u = r_i^v = 1$) without violating the constraints, i.e. tiles $u$ and $v$ do not contain the same column ($c_j^u c_j^v = 0 \ \forall\ j$), and 0 otherwise. $D$ can be written as $D = \big[CC' = 0\big]$ where Iverson's notation is applied element-wise on the matrix. The search space of tiles which can contain the same row while satisfying the constraints, is then reduced to the cliques of the graph with $D$ as its adjacency matrix. A generous upper bound on the number of cliques that need to be evaluated is $\mathcal{O}(2^T)$. In practice, due to the interplay between rows and columns, the search space is significantly restricted. The same procedure can be performed to optimize the columns.

# 4 EXPERIMENTAL EVALUATION

We compared MTA-SP and MTA-ICM to two additional methods that can be used to discover tiles. The plaid model (Lazzeroni & Owen, 2002) finds overlapping tiles, or 'layers' in the data one at a time, in a greedy fashion so that each consecutive tile $t$ is searched for in the residual left over from the previous $t-1$ tiles. For each tile, we determined whether it accounted for the background model or for the data and evaluated it accordingly. Another approach to MTA is to quantize the output of a matrix factorization technique. We experimented with a technique based on PCA (Jolliffe 1986).[4]

### 4.1 SYNTHETIC DATA

For the purpose of evaluating the different techniques we generated ~2000 synthetic data matrices of various sizes, each containing several tiles with varying tile dimensions. We analyzed the matrices using MTA-SP, MTA-ICM, the plaid method, and PCA, and evaluated their ability to correctly identify tiles. Each matrix in the data set is an $N \times N$ matrix, with $N = 40, 70, 100, 150, 200$. The number of tiles in each matrix ranges from 1 to 10. The average area of the tile is held constant across all the matrices and each tile covers approximately 4% of the data matrix. The tiles were randomly generated, subject to the constraint that two tiles cannot contain the same matrix element. The data was corrupted with additive Gaussian noise with $\log_{10}(\sigma^2)$ =-1.5, -1.0,-0.8,-0.675,-0.55,-0.425, and -0.3. Fig. 1 shows an example of a data matrix before adding noise, and shows a possible solution for the data after applying MTA to the *corrupted* data (not shown), where rows and columns have been reordered based on the analysis output to reveal the tiles. For each setting of matrix size, number of tiles, and noise level, we generated 20 different matrices, on which we tested the performance of all algorithms, except the plaid method which was only tested between 1-10 times for each setting due to the user intensive nature of the plaid software[5] and its inability to run in batch mode.

The input to the algorithms was set to be the log likelihood ratio of each element being in a tile versus being part of the background, calculated by assuming a normal distribution and fixing the standard deviation to 0.5 regardless of the true noise level. Additionally PCA and the plaid method take as input the noisy data matrix. The output of each algorithm is an integer value matrix where 0 indicates an element is part of the background model, and non-zero values indicate the tile index. Model selection (*i.e.* determining the correct number of tiles) was performed by evaluating the following cost function

$$\sum_{t,i,j} r_i^t c_j^t \log \frac{\ell_{ij}^0}{\ell_{ij}} - \sum_{i,j} \log \ell_{ij}^0 + T(N+M)\log 2. \quad (10)$$

[4]To find $T$ tiles, the $N \times N$ covariance matrix of the columns of $X$ is computed and its $T$ principal components are extracted (this procedure can arbitrarily be applied to the rows instead of the columns). Since each principal component corresponds to a major direction of variation across the columns, the values in each component provide evidence for the rows belonging to a corresponding tile. For each component, the rows belonging to the tile are identified by comparing the elements in the component to a threshold. To obtain maximum separation of values in the two groups, the threshold is selected so as to minimize the derivative of the cumulative distribution of the elements. Each component is thresholded separately to produce a set of $T$ binary vectors, each corresponding to a tile. Next, each column of the data matrix is robustly projected onto these vectors using a pseudo-inverse to account for non-orthogonalities introduced by thresholding. This yields a set of $T$ real numbers for each row of data, which are thresholded to determine the set of tiles that the row belongs to.

[5]www-stat.stanford.edu/~owen/clickwrap/plaid.html

The cost function is the minimum description length (MDL) for coding the model, assuming the distribution over the row and column indicator variables is uniform (which is strictly not true because of the term $\frac{1}{h(T)}$ in (2)). The number of tiles $T$ the algorithm was allowed to discover was incremented until it incurred an increase in the cost function. Since the plaid method produces overlapping tiles, some of which may model the background, overlapping elements of tile identification was resolved under this cost function to produce sensible results.

Performance was evaluated under several different criteria. First, we computed the Hamming distance between the 'ground truth' and the matrix output by each algorithm (after setting any non-zero values to 1 in the output matrix and comparing it to the clean input matrix) and normalizing by the size of the matrix. The Hamming distance gives a general measure of how well the algorithm was able to identify which elements in the input matrix belong to tiles and which belong to the background model. Fig. 3 shows the results for a set of experiments evaluated at different dimensions of an operating point chosen to be a matrix size of $100 \times 100$ and 5 tiles. We show results where the matrix size is fixed and the number of tiles is varied, and vice versa. All noise levels are shown for each setting of matrix size and number of tiles. The plots are shown with a log scale on both axes, and each point is an average of 20 experiments. Note that the scales in the y-axis differ across plots. The Hamming distance shows that MTA-SP performs better than the other algorithms, for most experimental settings. MTA-SP and MTA-ICM perform very similarly for large number of tiles. PCA and the plaid method both perform poorly, for the most part due to their inability to identify non-background elements. The increase in the number of tiles results in a decrease in performance while increasing the matrix size does not effect the performance significantly. Changes in the noise level have a noticeable effect only for a small number of tiles, and as would be expected, all algorithms succeed at finding a single tile.

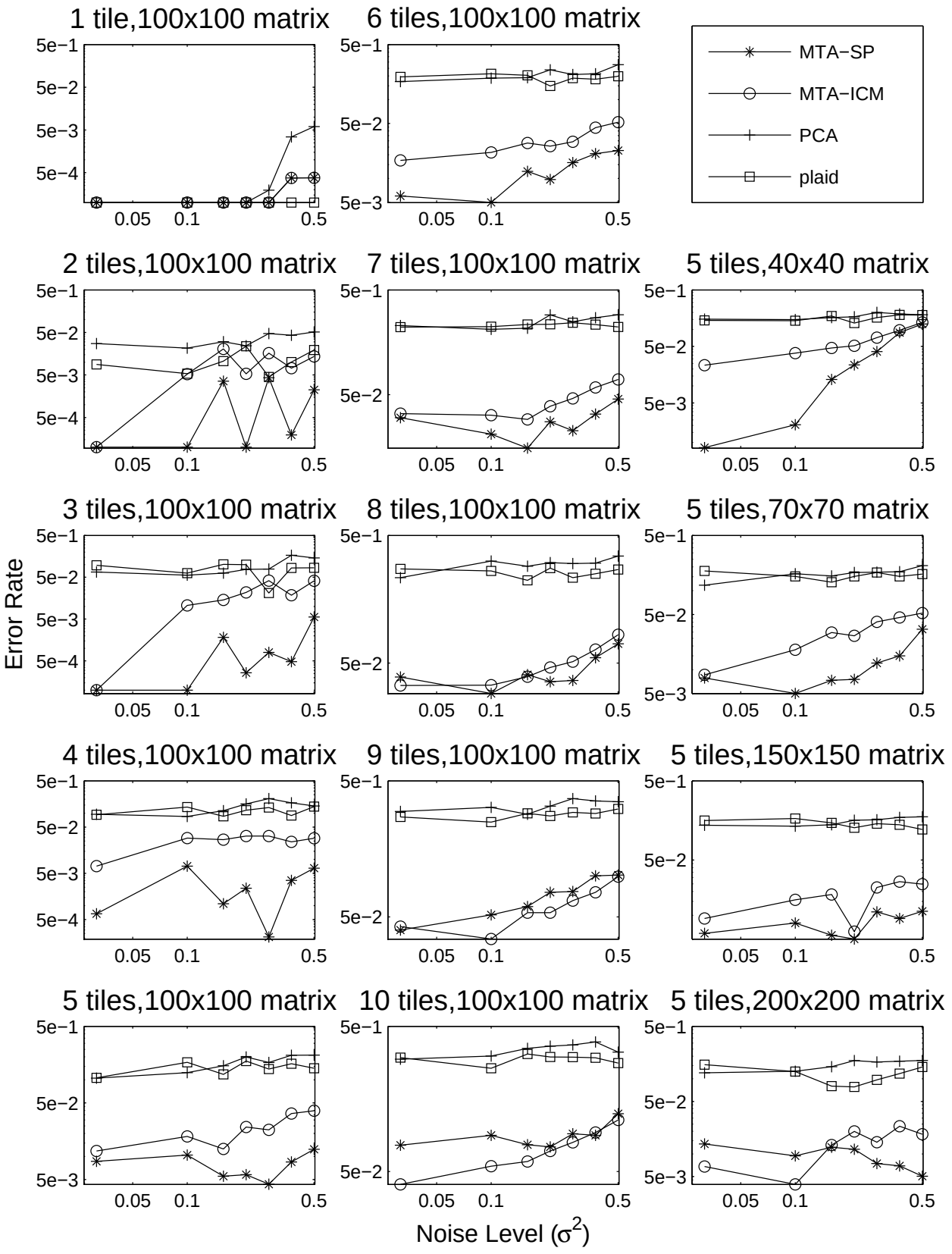

Figure 3: Hamming distance for different settings of matrix size and number of tiles.

Hamming distance does not provide information about how well the algorithm was able to identify the actual tiles that make up the non-background elements of the matrix. We address this question by evaluating the classification error made by the different algorithms. This evaluation requires that the tiles discovered by the algorithms be matched to the 'ground truth', which while used to generate the data, is not necessarily the best tile configuration. Tiles in the 'ground truth' were matched to the experimentally determined tiles to which they most overlapped. The matching was performed in a greedy fashion such that at most, a one-to-one correspondence was established between tiles. The tiles in the matrix output were then re-numbered based on this matching. The classification error metric is simply the number of mismatches between the matrices normalized by the size of the matrix. The classification errors are shown in Fig. 4 for the same set of experiments as previously discussed.

MTA-SP outperforms the other methods in all but 4 of the experiments. In the cases where MTA-ICM does better in terms of the Hamming distance, we see it suffers from more classification errors. This indicates that for these cases, MTA-ICM correctly identified more elements to be in tiles and in the background, but it did so with less accurate tiling with regard to the actual data. Overall, both MTA-SP and MTA-ICM perform considerably better than the plaid method and PCA which struggle to identify tile elements.

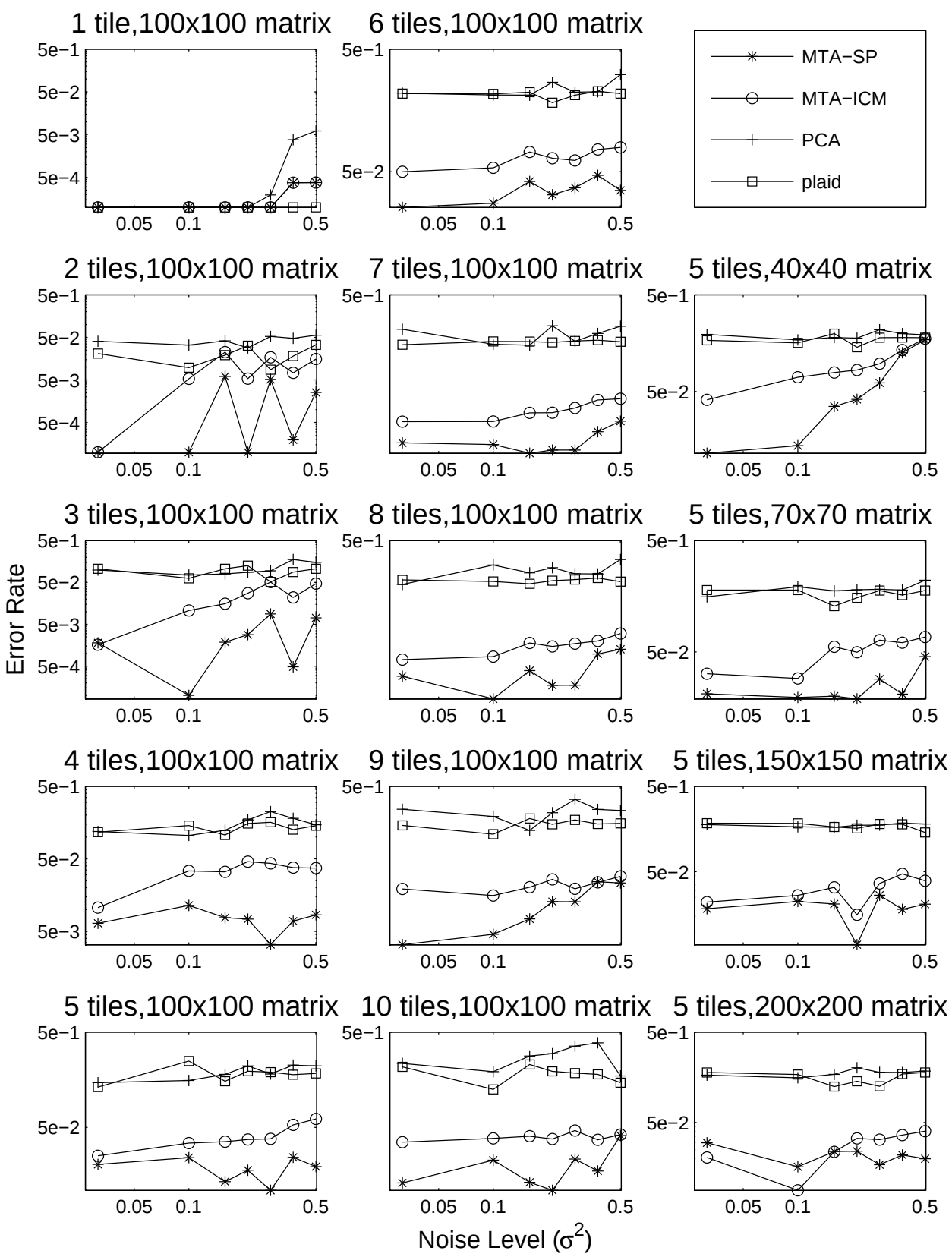

Figure 4: Classification error for different settings of matrix size (right column) and number of tiles.

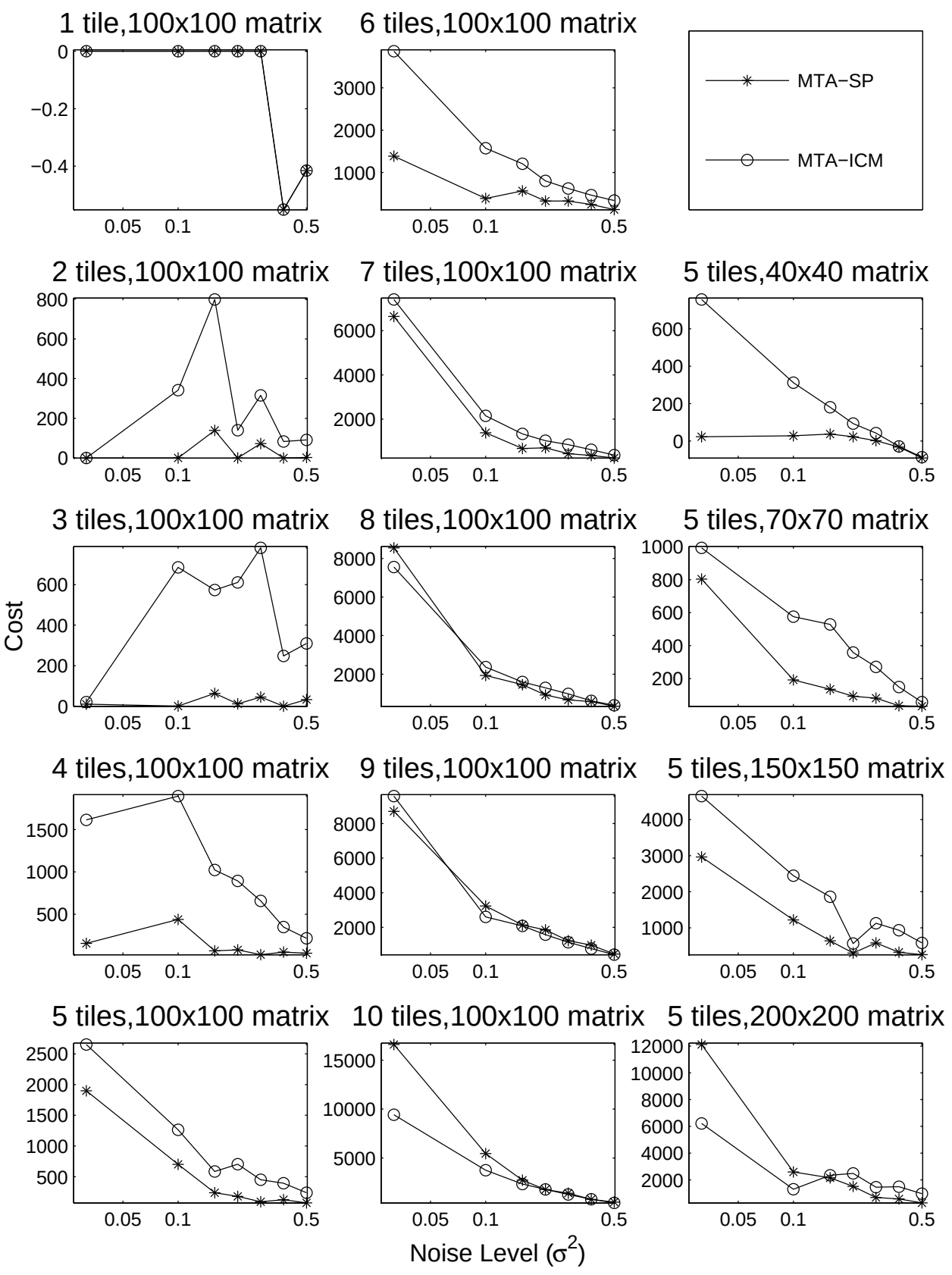

Figure 5: Cost function relative to ground truth for different matrix sizes and numbers of tiles.

Fig. 5 compares the performance of MTA-SP and MTA-ICM in terms of the cost function in (10) they both minimize. The other algorithms are not shown as they perform several orders of magnitude worse, a consequence of the fact they do not directly minimize the cost function. The values are shown after subtraction of the cost of the 'ground truth'. For the cases where the algorithms performed perfectly, the incurred cost was zero. As the complexity of the problem increases, so does the cost incurred. This is not directly due to the added component of $T(N+M)$ which appears in the 'ground truth' model as well, though it may contribute towards the higher costs if the algorithms find more tiles then necessary. There are cases where the algorithm performs better than the true model (e.g., 5 tiles, $40 \times 40$). This phenomena occurs because the noise applied is such that some of the rows and/or columns of a tile are better modelled in the base model than as part of a tile, or vice versa. However, under the 'ground truth', the tile is still evaluated as it was originally generated, despite the amount of corruption by the noise. In general, the algorithms do not reach the global minimum and have a higher cost than that of the ground truth. We see the same trends as before, with MTA-SP doing better on average, and having only a small difference when it fails to do as well as MTA-ICM.

## 4.2 YEAST GENETIC INTERACTIONS

The 'synthetic genetic array' data set (SGA) (Tong *et al.*, 2004) describes events of lethality for double knock-out genes in yeast, where the single knock-out strains are viable. The data set is a binary matrix of $135 \times 1023$ (Fig. 6), where a value of 1 at position $x_{ij}$ denotes a synthetic genetic interaction, i.e. the double mutant whose genes $i$ and $j$ have been knocked out is not viable and thus both genes are required for survival, although either one by itself is dispensable.

We compared the performance of the different methods by obtaining clusters of genes from each method and analyzing their biological significance in terms of higher than expected rates of similarly annotated genes, using Gene Ontology annotations[6]. For each cluster, we evaluated the $p$-value of the Bonferroni-corrected hyper-geometric distribution for observing the number of genes with a specific annotation. In order to compare MTA to different types of analysis, we obtained clusters from hierarchical agglomerative clustering (HAC), the original plaid model, and MTA. We allowed up to 15 clusters, where for HAC the dendrogram was used to obtain row/column clusters, and for MTA, we considered the genes corresponding to row elements of a tile as elements of the row cluster and genes corresponding to column elements of a tile as column clusters. MTA identified more statistically significant clusters ($p < 0.01$) for different GO-category aspects (cellular component, molecular function and biological process) than either HAC, or the unmodified plaid model. Overall, MTA identified 17 significant groups, HAC identified 10, and the plaid method identified 5 (see Table 1).

Table 1: Number of times groups of yeast genes are significantly enriched for Gene Ontology categories

| Biological Aspect | MTA | HAC | Plaid |
|---|---|---|---|
| Biological Process | 7 | 4 | 4 |
| Cellular Component | 8 | 4 | 1 |
| Molecular Function | 2 | 2 | 0 |
| Total | 17 | 10 | 5 |

## 5 DISCUSSION

We introduced a new analysis framework, matrix tile analysis, in which an input matrix is decomposed into a set of non-overlapping components, or tiles. Unlike many models for matrix decomposition, our method does not require addition and multiplication to be defined over data vectors, and unlike approaches such as bi-clustering, different components are allowed to overlap in the rows and columns, though not in both simultaneously. We described a probabilistic model of the problem and its graphical representation using a factor-graph. We compared the performance of four different approaches for solving MTA and were able to demonstrate that the sum-product algorithm (MTA-SP) performs better on a large set of synthetic problems compared to a greedy algorithm, the plaid method, and PCA. We were also able to extract biologically related groups of genes by applying MTA to yeast genetic interaction data. Tiles identified by MTA in two other gene interaction data sets are being biologically verified.

Our model requires that the input be given in terms of data and background likelihoods. This was motivated by data for which prior analysis or expert knowledge resulted in likelihood measurements. However, for the general case, in the absence of strong prior knowledge about the data distribution, it is always possible to compute likelihood estimates from empirical measurements. Further, because of the formulation of the model under a probabilistic framework, the model can be extended to jointly infer the tiling and learn the parameters in the likelihood functions, using the EM algorithm.

[6] ftp://genome-ftp.stanford.edu/pub/yeast/

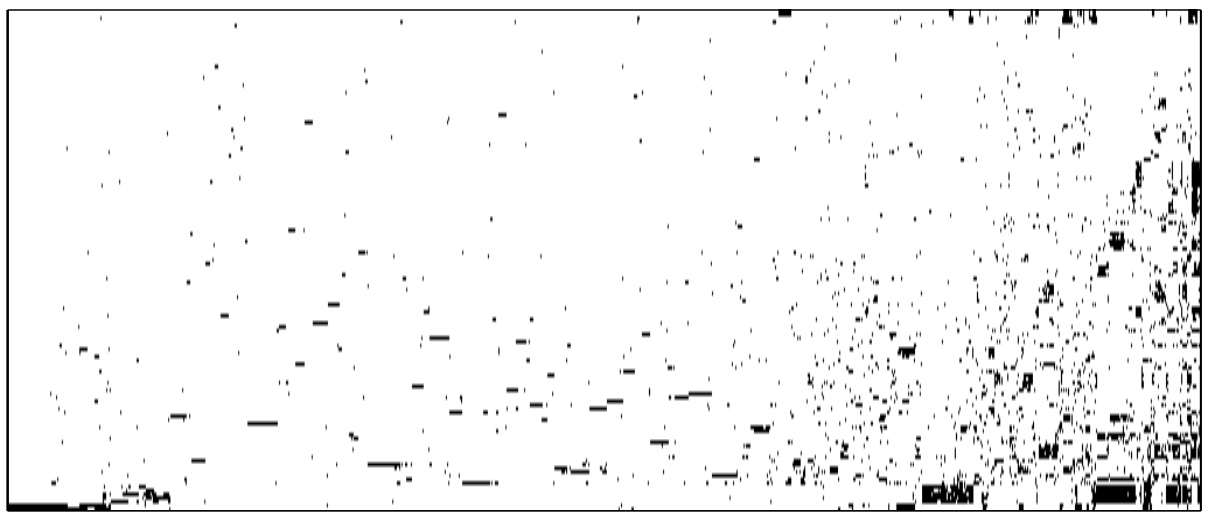

Figure 6: Yeast genetic interaction data. The rows and columns are ordered using clustering.

### References

EM Airoldi, *et al.* A Latent Mixed Memebership Model for Relational Data. *Proc ACM Link-KDD Workshop*, 2005.

AJ Bell, TJ Sejnowski. An information maximization approach to blind separation and blind deconvolution. *Neur Comp* **7**, 1129–1159, 1995.

Y Cheng, GM Church. Biclustering of expression data. *Proc Int Conf Intell Syst Mol Biol* **8**, 93-103, 2000.

D Dueck, *et al.* Multi-way clustering of microarray data using probabilistic sparse matrix factorization. *Proc Int Conf Intell Syst Mol Biol* **13**, 2005.

MV Eisen, *et al.* Cluster analysis and display of genome-wide expression patterns. *Proc of the National Academy of Sciences* **95**, 14863–14868, 1999.

IT Jolliffe. Principal Component Analysis. Springer-Verlag, New York NY, 1986.

FR Kschischang, *et al.* Factor graphs and the sum-product algorithm. *IEEE Trans Info Theory* **47:2**, 498–519, 2001.

L Lazzeroni, A Owen. Plaid Models for Gene Expression Data. *Statistica Sinica* **12:1**, 61-86, 2002.

D Lee, S Seung. Learning the parts of objects by non-negative matrix factorization. *Nature* **401**, 788–791, 1999.

P Resnick, *et al.* GroupLens: An Open Architecture for Collaborative Filtering of Netnews. *Proc ACM Conf Comp Supp Coop Work*, 175-186, Chapel Hill, NC, 1994.

N Srebro, *et al.* Maximum-Margin Matrix Factorization. *Advances in Neural Info Proc Sys* **17**, 2005.

AH Tong, *et al.* Global Mapping of the Yeast Genetic Interaction Network. *Science* **303**, 808–813, 2004.